\documentclass[preprint,11pt]{elsarticle}

\usepackage[nohyperlinks]{acronym}
\usepackage{microtype}
\usepackage{hyperref}
\usepackage{orcidlink}
\usepackage{xcolor}
\usepackage{graphicx}
\usepackage{listings}
\usepackage{upquote}
\usepackage{tikz}
\usepackage{etoolbox}
\usepackage{booktabs}

\hypersetup{
	colorlinks=true,
	linkcolor=blue,
	citecolor=blue,
	urlcolor=blue,
}

\newcommand{\bi}{\begin{itemize} \itemsep -2pt}
\newcommand{\ei}{\end{itemize}}

\newcommand{\fig}[1]{Fig.~\ref{#1}}
\newcommand{\mytab}[1]{Tab.~\ref{#1}}

\definecolor{deepblack}{rgb}{0,0,0}

\newcommand\plainlststyle{\lstset{
language=Python,
showstringspaces=false,
basicstyle=\ttfamily\scriptsize,
otherkeywords={self},
keywordstyle=\ttfamily\scriptsize,
emphstyle=\ttfamily\scriptsize,
stringstyle=\color{deepblack},
commentstyle=\color{deepblack},
frame=tb,
numbers=none,
breaklines=true,
numberstyle=\ttfamily\scriptsize,
columns=flexible,
}}

\newcommand{\tikzpicAnnotate}[4]{
{}{}{}
\begin{tikzpicture}
	\node[above right, inner sep=0] (image) at (0,0) {
    \includegraphics[width=\columnwidth]{#1}
};
\node [anchor=north, xshift=#3, yshift=#4] at ([yshift=-2pt]image.south) {#2};
\end{tikzpicture}
}

\newcommand{\minipageFig}[3]{
\begin{minipage}[b]{#1\columnwidth}
\tikzpicAnnotate{#2}{#3}{0}{0}
\end{minipage}
}

\acrodef{AI}{Artificial Intelligence}
\acrodef{CNN}{Convolutional Neural Network}
\acrodef{GPT}{Generative Pre-trained Transformer}
\acrodef{IR}{Intermediate Representation}
\acrodef{LLM}[LLM]{Large Language Model}
\acrodef{NLP}{Natural Language Processing}
\acrodef{NODE}{Neural Ordinary Differential Equation}
\acrodef{PANN}{Physics Augmented Neural Network}
\acrodef{PINN}{Physics-Informed Neural Network}
\acrodef{RLHF}{Reinforcement Learning from Human Feedback}
\acrodef{ViT}{Vision Transformer}
\acrodef{VLM}{Vision Language Model}
\acrodef{PP}{Percentage Point}

\definecolor{colM0}{RGB}{100,100,100}
\definecolor{colM1}{RGB}{31,119,180}
\definecolor{colM2}{RGB}{214,39,40}
\definecolor{colM3}{RGB}{44,160,44}
\definecolor{colM4}{RGB}{255,127,14}

\newcommand{\circnum}[1]{%
\tikz[baseline=(char.base)]{
\node[
    shape=circle,
    draw,
    inner sep=1pt
] (char) {\small #1};
}%
}

\newcommand{\symbolBaseline}{\textcolor{colM0}{\textsf{\circnum{0}}}}
\newcommand{\symbolNaive}{\textcolor{colM1}{\textsf{\circnum{1}}}}
\newcommand{\symbolNaiveText}{\textcolor{colM2}{\textsf{\circnum{2}}}}
\newcommand{\symbolIR}{\textcolor{colM3}{\textsf{\circnum{3}}}}
\newcommand{\symbolIRText}{\textcolor{colM4}{\textsf{\circnum{4}}}}

\makeatletter
\patchcmd{\pprintMaketitle}{\elsaddress\par\vskip36pt}{\elsaddress\par\vskip12pt}{}%
  {\PackageWarningNoLine{main_EML}{Could not reduce the gap between affiliations and abstract}}
\patchcmd{\MaketitleBox}{\elsaddress\par\vskip36pt}{\elsaddress\par\vskip12pt}{}{}
\makeatother

\journal{}

\begin{document}

\begin{frontmatter}

\title{Perception Before Simulation: Sketch-Based Generation of Multibody Simulation Code With Vision Language Models}

\author[innsbruck]{Tobias\ Möltner\,\orcidlink{0009-0004-2471-9439}
\footnote{Corresponding author E-mail: \href{mailto:tobias.moeltner@uibk.ac.at}{tobias.moeltner@uibk.ac.at}}}
\author[augsburg]{Peter\ Manzl\,\orcidlink{0000-0002-7707-5524}}
\author[innsbruck]{Sebastian\ Weyrer\,\orcidlink{0009-0006-7898-4208}}
\author[innsbruck]{Michael\ Pieber\,\orcidlink{0000-0002-3745-3439}}

\affiliation[innsbruck]{organization={Department of Mechatronics, University of Innsbruck},
            addressline={Technikerstraße 13},
            city={Innsbruck},
            postcode={6020},
            country={Austria}}

\affiliation[augsburg]{organization={Institute of Materials Resource Management (MRM), University of Augsburg},
            addressline={Am Technologiezentrum 8},
            city={Augsburg},
            postcode={86159},
            country={Germany}}
\begin{abstract}
Large language models (LLMs) can generate multibody simulation models from textual descriptions, but for complex mechanical systems these descriptions become lengthy and error-prone, whereas a sketch conveys the same mechanical system compactly. This study investigates whether Vision Language Models (VLMs) can generate physically correct multibody simulation code directly from sketches of mechanical models. In an automated closed-loop benchmark, 23 parameterized mechanical models are rendered as annotated sketches, and every generated simulation model is verified against ground truth by executability, graph isomorphism, and parameter and trajectory agreement. Across 12 open-weight VLMs, a two-stage pipeline that first extracts a code-agnostic intermediate representation of the sketched mechanical model doubles the success rate of simulation code generation compared to direct code generation by the VLM ($48.0\,\%$ vs.\ $23.9\,\%$). With code generation held fixed by a single coder LLM, so that only perception varies, replacing the textual description with a sketch lowers success from $94.8\,\%$ to $48.0\,\%$, and to $78.2\,\%$ when the textual description accompanies the sketch. An ablation attributes the gain to pre-selecting the element types shown in the sketch. Error classification and robustness tests to scaling and rotation locate the remaining failures in reading the drawing rather than in code generation. The pipeline transfers to hand-drawn sketches. Turning an engineer's sketch directly into a verified dynamic simulation opens a route to automating model creation in computational mechanics.
\end{abstract}


\end{frontmatter}

\section{Introduction}
Data-driven surrogates such as \acp{PINN}~\cite{cai2021PINNS_heat}, \acp{PANN}~\cite{klein2022PANN}, \acp{NODE}~\cite{chen2018neuralODEs, wang2025mbd} or, for damped multibody systems, SLIDE~\cite{manzl2025SLIDE} can directly predict the response of mechanical systems, but they rely on large datasets and offline training and must be retrained for new problems. Explainable AI~\cite{rai2020explainableAI} can reveal part of their inner workings. White-box simulation models avoid this, yet typically require manual design and comprehensive system knowledge~\cite{yang2017investigatingGreyBoxModeling}. A multibody simulation model, for example, describes a mechanical system as bodies connected by joints, springs and constraints and loaded by forces, from which the solver assembles the equations of motion. \acp{LLM} offer a way to create such simulation models automatically as code, as our previous work showed for multibody simulation models generated from text~\cite{gerstmayr2024multibodyModelsGenerated, Moeltner2026_CreationEvaluationSelfValidation}, including the MecEng benchmark~\cite{Gerstmayr2026_LLMAwarenesMechanics} (Sec.~\ref{sec_sota}).

All of these studies use textual descriptions of mechanical models as input. As systems grow, such descriptions become lengthy and error-prone, while engineers naturally communicate a mechanical system as a sketch (\fig{fig:exampleSketches}). MecEng's outlook accordingly proposes a preceding module that turns images or sketches into accurate descriptions~\cite{Gerstmayr2026_LLMAwarenesMechanics}. This work builds and evaluates that module. It keeps MecEng's evaluation and changes one variable, the input modality.

\noindent The contributions of this work are:
\bi
\item a quantification of the text-to-sketch gap for multibody simulation model generation across 12 open-weight \acp{VLM}, with text, sketch-plus-text and sketch-only conditions on identical tasks,
\item a perception-first pipeline, in which a closed-vocabulary element pre-selection feeds a code-agnostic \ac{IR} that a separate coder \ac{LLM} turns into simulation code, together with an ablation that locates its benefit, and
\item a robustness and transfer analysis covering scaled, rotated and hand-drawn sketches, and an error classification that localizes the failures.
\ei

\section{Related work}\label{sec_sota}
\paragraph{Text to simulation} In multibody dynamics, two strategies have been used to generate physics-based simulation code with \acp{LLM}. Wang et al.~\cite{wang2025Chronollm} customize \acp{LLM} for the multibody code PyChrono, mainly by continued pre-training and fine-tuning. The associated benchmark SimBench~\cite{wang2026simbench} evaluates the generated PyChrono scripts in multi-turn interactions with a simulator-specific \ac{LLM} as judge, which scores them against predefined rules using reference code and documentation. Our previous work instead uses \acp{LLM} without fine-tuning and provides the required knowledge by in-context learning with retrieved context for the multibody code Exudyn~\cite{Exudyn2024}. Gerstmayr et al.~\cite{gerstmayr2024multibodyModelsGenerated} showed that frontier \acp{LLM} can generate multibody simulation models from natural language descriptions. Previous work~\cite{Moeltner2026_CreationEvaluationSelfValidation} developed a parameterized virtual laboratory that creates, evaluates and self-validates such simulation models with a broad range of open-weight \acp{LLM}. MecEng~\cite{Gerstmayr2026_LLMAwarenesMechanics} extends this to flexible multibody systems with exact 3D geometry, finite-element meshing and model order reduction, and separates structural from parameter errors by system-graph isomorphism, all compared deterministically against expert ground truth. All of these approaches take text as input. The present study adopts the component selection and code generation of~\cite{Moeltner2026_CreationEvaluationSelfValidation} and the graph-based checks of MecEng, and replaces the text by a sketch.

\paragraph{Vision language models} \acp{LLM}, built on the transformer~\cite{2017_Vaswani_attentionIsAllYouNeed, 2019_Devlin_BERT_PreTrainingDeepBidirectionalTransformersForLanguagaeUnderstanding, Lewis_2020_BART, 2018_radford_improvingLanguageUnderstandingGPT, 2019_Radford_languageModelsUnsupervisedMultitaskLearners, 2020_Brown_languageModelsAreFew-ShotLearners, yang2025qwen3} and improved by scaling~\cite{2020_Kaplan_scalingLawsForNeuralLanguageModels} and alignment~\cite{ouyang2022trainingLLMsRLHF}, generalize to programming~\cite{zhuo2025bigcodebench, jimenez2024SWEBench, Singh2025GPT5ModelCard, Hui2024qwen2_5Coder, Guo2024deepseekCoder_LLMMeetsProgramming}, although whether this reflects genuine understanding is debated~\cite{bender2021dangers, hendrycks2020MMLU, bubeck2023paperSparksOfAGI}. Vision transformers~\cite{dosovitskiy2020VisionTransformer, radford2021learningTransferableVisualModelsFromNLPSuperv, Caron_2021_SelfSupervisedViT, He2022maskedAutoencoders_ViT} provide the image encoders of modern \acp{VLM}. \acp{VLM} extend \acp{LLM} to images and, unlike detectors with a fixed set of categories~\cite{redmon_2016_YOLO_original}, reason with open vocabularies over visual and textual input~\cite{teamQwen2023qwenVL}. LLaVA~\cite{Liu2023Llava} connected a pre-trained vision encoder to an \ac{LLM} through visual instruction tuning, and GPT-4V~\cite{openai2023gpt4v} demonstrated this at the scale of frontier \acp{LLM}. Among open-weight \acp{VLM}, Qwen2-VL~\cite{wang2024qwen2VL} processes images at their native resolution, which matters for detail-rich inputs such as technical drawings, and Qwen2.5-VL~\cite{bai2025qwen25vl} and Qwen3-VL~\cite{bai2025qwen3VL} extend this line. This study also evaluates the \acp{VLM} Gemma~3~\cite{kamath2025gemma3}, Gemma~4~\cite{gemmateam2026gemma4} and Ministral~3~\cite{mistralai2025ministral3}. Benchmarks relevant to sketch interpretation address mathematical reasoning from images~\cite{wang2024MathVisionBenchmark}, translating visual designs into code~\cite{Si2025Design2Code}, multi-step spatial reasoning~\cite{Zeng2025LEGOpuzzles} and intrinsic geometric understanding~\cite{Abdullaeva2026NoReGeo}, and survey spatial-reasoning deficits of \acp{LLM}~\cite{Sharma2023SpatialReasoningLLM, Yamada2024SpatialLLM}.

\paragraph{Large language models for engineering artifacts} Text2CAD~\cite{khan2024text2cad} and CAD-Coder~\cite{guan2026cad} generate parametric CAD from text, and sketch2prototype~\cite{edwards2024sketch2prototype} turns a single hand-drawn sketch into a 3D-printable prototype. For simulation beyond multibody dynamics, FEABench~\cite{mudur2025feabench} evaluates whether \acp{LLM} and \ac{LLM} agents solve multiphysics problems from natural-language descriptions by operating the finite-element software COMSOL Multiphysics, and MetaOpenFOAM~\cite{chen2024metaopenfoam} uses multiple \ac{LLM} agents to set up and run computational fluid dynamics simulations in OpenFOAM. OpenSeeSimE~\cite{Ezemba2026OpenSeeSimE} benchmarks \ac{VLM} question answering on engineering simulation scenes, and EngiAI~\cite{molinari2026engiai} and EngiBench~\cite{Zhou2025EngiBench} evaluate \acp{LLM} across broader engineering workflows.\\

\noindent None of these works generates executable dynamic simulation models from drawings and verifies them against ground truth, which is the gap addressed here.

\section{Methods}\label{sec_Methods}
\subsection{Benchmark and sketches}\label{sec_benchmark}
\begin{figure}[t]%
\centering
\minipageFig{0.55}{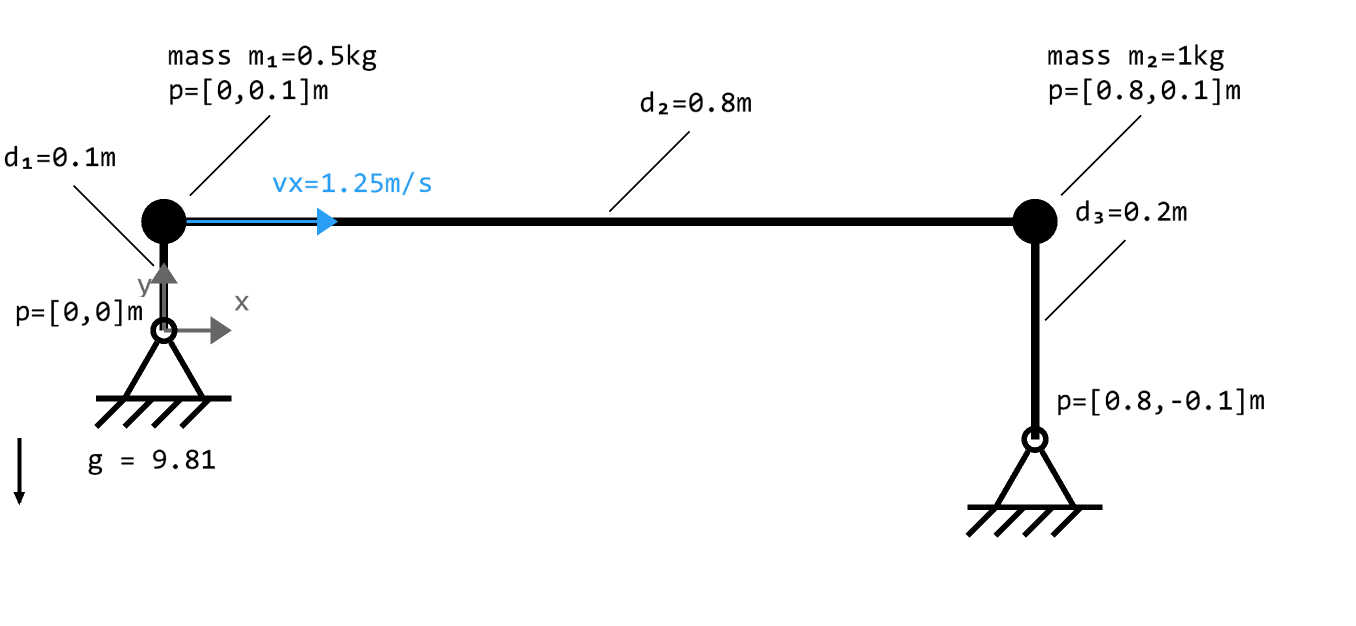}{a)}
\minipageFig{0.42}{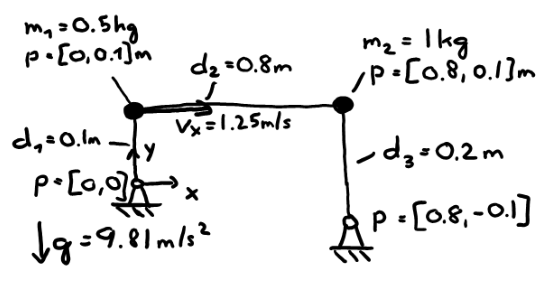}{b)}
\caption{Four-bar mechanism (ID 19, \mytab{tab:MBSmodels}) with point masses and distance constraints: a) generated sketch, b) hand-drawn sketch of set A.}%
\label{fig:exampleSketches}%
\end{figure}

The benchmark comprises 23 mechanical models (\mytab{tab:MBSmodels}), from single mass points over pendulums, oscillators, rigid bodies, and a four-bar mechanism to a spring truss and three beam structures. For each mechanical model, a domain expert implemented a parameterized ground-truth simulation model in Exudyn~\cite{Exudyn2024}\footnote{Version 1.11.0.}, which is instantiated at 5 randomly drawn parameter sets (115 tasks). The parameter sets vary, e.g., masses, forces, distances and the number of bodies. Each instance is rendered as an annotated sketch by visualizing its ground-truth simulation model in a sketch-like style (\fig{fig:exampleSketches}a). The sketch shows every element with the information needed to build the simulation model, since missing or ambiguous information would propagate into the evaluation. Objects and the points of application of forces, torques and constraints have fixed positions, annotations float to avoid overlaps, and sine and cosine are shown together with angles to avoid round-off.

Every sketch is standardized to $1092\times728$\,pixels (aspect ratio 1.5), multiples of 28\,pixels, which is one visual token of the Qwen2.5-VL encoder (14-pixel patches merged $2\times2$). Each \ac{VLM} then applies its own image preprocessing. These 115 standard sketches are used for all results unless stated otherwise. To test robustness, the standard sketches are down-scaled from $0.9\times$ to $0.2\times$ in steps of 0.05 and rotated within $\alpha=\pm45^\circ$ in steps of $5^\circ$, re-padded to the same canvas without cutting off content, and evaluated for four \acp{VLM} of two model families (Qwen and Gemma) that span the performance range. Rotations are applied symmetrically, so that the agreement between $+\alpha$ and $-\alpha$ distinguishes an orientation effect from noise. To test genuine drawings, 8 of the mechanical models (IDs 1, 3, 4, 6, 7, 8, 11 and 19) were drawn on a tablet by three people (\fig{fig:exampleSketches}b, sets A, B and C). Set A was used during pipeline development and is not held out, and sets B and C were redrawn from the set-A images, so the three sets share layout and parameters and differ in handwriting, notation and annotation style.

\begin{table}[t]%
\caption{Mechanical models of the benchmark, numbered by increasing complexity.}
\label{tab:MBSmodels}
\centering
\resizebox{\linewidth}{!}{
\begin{tabular}{rl@{\hspace{1.5em}}rl}
\toprule
ID & model name & ID & model name \\
\midrule
1 & flying mass point & 13 & n-pendulum \\
2 & free fall mass point & 14 & n-pendulum elastic spring \\
3 & single mass oscillator & 15 & mass point on string elastic \\
4 & double mass oscillator & 16 & mass point on string rigid \\
5 & single mass oscillator gravity & 17 & flying rigid body \\
6 & two mass points with distances & 18 & double pendulum rigid bodies \\
7 & two mass points with springs & 19 & four bar mechanism point masses \\
8 & single pendulum elastic spring & 20 & springs bridge \\
9 & double pendulum & 21 & beam example \\
10 & double pendulum elastic spring & 22 & beam cantilever \\
11 & n-mass oscillator & 23 & beam bridge \\
12 & single pendulum & & \\
\bottomrule
\end{tabular}
}
\end{table}

\subsection{Pipelines}\label{sec_pipelines}
\begin{figure}[t]%
\includegraphics[width=\columnwidth]{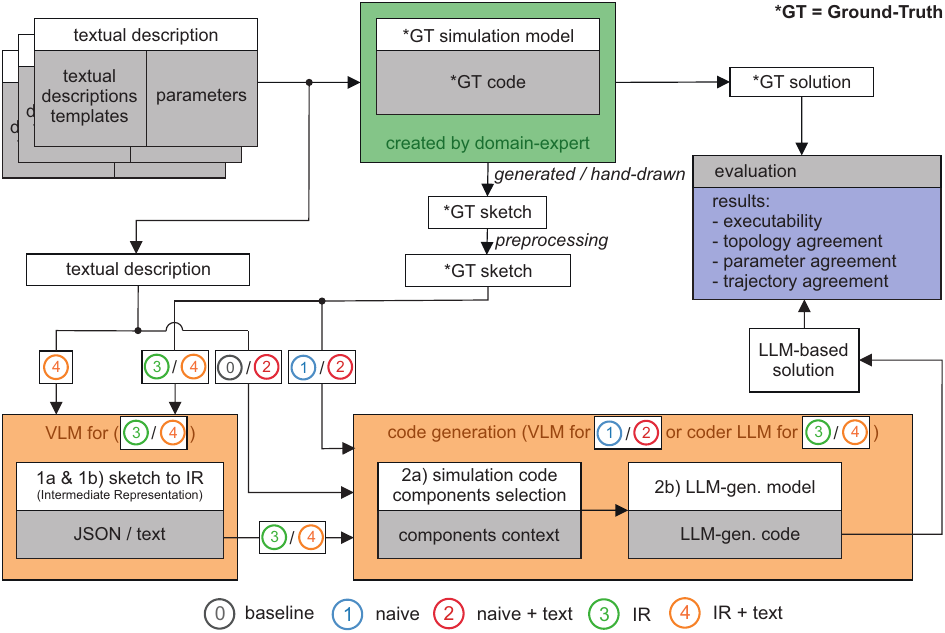}%
\caption{Pipeline for code generation from sketches. Sketches are rendered from parameterized ground-truth simulation models, converted into simulation code by one of the pipelines of \mytab{tab:pipelineVersions}, and evaluated against the ground truth.}%
\label{fig:pipeline}%
\end{figure}

\fig{fig:pipeline} summarizes the process from the ground-truth simulation model to the evaluation. A pipeline run consists of up to four stages (\mytab{tab:pipelineVersions}). Stages 1a and 1b extract the \ac{IR} from the sketch (Sec.~\ref{sec_IR}). Since \acp{LLM} can fail at seemingly simple arithmetic~\cite{qian2023limitationsofLLMsArithmetics}, they are used to write code for a simulation tool rather than to compute results directly. Exudyn models are written in Python, which is widely represented in the training data of \acp{LLM}. Accordingly, only 2 of the 5\,635 codes generated for the standard sketches contain Python syntax errors. Stages 2a and 2b follow the code-generation approach of~\cite{Moeltner2026_CreationEvaluationSelfValidation}. In Stage 2a, the \ac{LLM} that writes the code selects the required Exudyn simulation components from a list of 27 available ones (e.g.\ ground, point mass, rigid bodies, spring-dampers, joints and 2D beams), and in Stage 2b it writes the simulation code, given the code syntax and usage of exactly these components as context. The \emph{naive} pipeline runs only Stages 2a and 2b, so the \ac{VLM} writes the code directly from the image, and \emph{naive + text} additionally receives the expert-written textual description of the task. \emph{IR} and \emph{IR + text} run all four stages, and a separate coder \ac{LLM} generates the code from the extracted \ac{IR}. Only \emph{naive} and \emph{IR} are deployable when a drawing is the only input. \emph{IR + text} extracts its \ac{IR} from sketch and text. In a text-only baseline, the coder \ac{LLM} receives the description instead of the sketch. For the flying mass point of \fig{fig:irExample}, the description reads:
\begin{quote}\itshape\small
Projectile motion of a point mass with the following properties: mass m = 10 kg, gravity g = 9.81 m/s\textsuperscript{2}, initial velocity in x/y/z-direction: vx = 5 m/s, vy = 1 m/s, vz = 0 m/s. The initial position is given as x=0 and y=0. Gravity acts along the negative y-axis, and there is no other external propulsion or resistance. The motion shall be evaluated for 1 s. Contact with ground is not modelled, so the projectile can go below y=0.
\end{quote}

The baseline and the \emph{IR} pipelines share the same fixed coder \ac{LLM}, Qwen2.5-Coder-32B-Instruct, the best-performing \ac{LLM} of~\cite{Moeltner2026_CreationEvaluationSelfValidation}. This isolates perception, because the input to the coder \ac{LLM} is the only difference between them. The \emph{naive} pipelines necessarily use the \ac{VLM} as coder \ac{LLM}, since the coder \ac{LLM} cannot process images. Both Stage-2 prompts are identical across all pipelines. Only the description of the mechanical model passed to them differs (none, the expert description or the extracted \ac{IR}), and in the \emph{naive} pipelines the sketch is attached as an image. All prompts are included in the published code (see Data availability).

\begin{table}[t]%
\caption{Pipeline variants. Where a coder \ac{LLM} input is listed, the code is written by the fixed coder \ac{LLM}, otherwise by the \ac{VLM}.}
\label{tab:pipelineVersions}
\centering
\begin{tabular}{rllll}
\toprule
& name & \ac{VLM} input & coder \ac{LLM} input & stages \\
\midrule
\symbolBaseline & baseline & -- & text & 2a, 2b \\
\symbolNaive & naive & sketch & -- & 2a, 2b \\
\symbolNaiveText & naive + text & sketch \& text & -- & 2a, 2b \\
\symbolIR & IR & sketch & \ac{IR} & 1a, 1b, 2a, 2b \\
\symbolIRText & IR + text & sketch \& text & \ac{IR} & 1a, 1b, 2a, 2b \\
\bottomrule
\end{tabular}
\end{table}

\subsection{Intermediate representation}\label{sec_IR}
\begin{figure}[t]%
\centering
\begin{minipage}[c]{0.40\columnwidth}
\centering
\includegraphics[width=\linewidth]{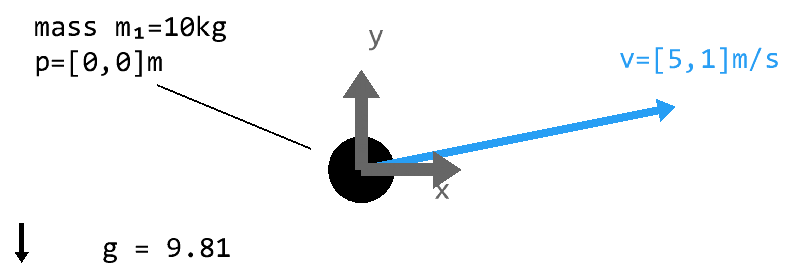}
\end{minipage}%
\hfill
\begin{minipage}[c]{0.56\columnwidth}
{\plainlststyle
\begin{lstlisting}
{
  "modelDescription": "...",
  "world": {"gravity":
    {"direction": [0.0,-9.81,0.0]}},
  "grounds": [],
  "parts": [
    {"id": "m1",
     "position": [0.0,0.0,0.0],
     "initialVelocity": [5.0,1.0,0.0],
     "mass": 10.0}
  ],
  "connectors": [], "constraints": [],
  "joints": [], "loads": [],
  "excitation": []
}
\end{lstlisting}
}
\end{minipage}
\caption{Sketch of the flying mass point (ID 1) and the JSON \ac{IR} extracted by Qwen3-VL-32B-Instruct (abbreviated).}
\label{fig:irExample}
\end{figure}

In Stage 1a, the \ac{VLM} selects the element types present in the sketch from a closed vocabulary of eleven tags: ground, point mass, spring damper, distance constraint, gravity, force, torque, rigid body, spherical joint, revolute joint and 2D beam. Each tag comes with a short description of its sketch convention, e.g.\ a zigzag line annotated with $k$ and/or $d$ for a spring damper and a plain bar for a distance constraint. The \ac{VLM} returns the selected tags as a JSON list. Only the \ac{IR} templates of the selected tags are passed to Stage 1b, which keeps only the relevant templates in the prompt. In Stage 1b, the \ac{VLM} writes the \ac{IR}, by default a JSON object that groups the elements into world (gravity), grounds, parts, connectors, constraints, joints, loads, beams and excitation (\fig{fig:irExample}). Besides its value, each field can carry a short free-text justification of how the \ac{VLM} read it from the sketch, e.g.\ why an arrow was interpreted as the gravity direction. \fig{fig:irExample} omits these justifications but shows the empty arrays of every generated \ac{IR}. The schema covers all mechanical models of the benchmark and is agnostic to the simulation code. It could equally target other code-based tools such as Project Chrono~\cite{TasoraNegrutChrono2013}. Two extraction rules encode explicit design decisions: a drawn gravity direction is taken as given, and a hinge circle counts as a spherical joint only if the sketch labels it so.

To locate the benefit of the \ac{IR}, two design choices are ablated on the 115 standard sketches for all 12 \acp{VLM}: JSON versus an equivalent free-text description (the same Stage-1b prompt, asking for prose instead of JSON), and with versus without Stage 1a, in which case the full representation is extracted in one step.

\subsection{Evaluation}\label{sec_evaluation}
Each generated simulation model is compared to its ground-truth simulation model with checks of MecEng~\cite{Gerstmayr2026_LLMAwarenesMechanics}, applied in the following order.

\emph{Executability.} Everything after \texttt{mbs.Assemble()}, the Exudyn call that finalizes the simulation model, is replaced by the solver call of the ground-truth simulation model, so that both simulation models are solved with identical settings. A generated simulation model is executable if it runs to completion under this call and writes a solution file.

\emph{Topology agreement or graph isomorphism.} The simulation model is converted into a graph whose nodes are its Exudyn items and whose edges are their mutual references. Items here can be Exudyn objects, nodes, markers, and loads. Sensors are excluded, as they do not affect the dynamics, and isolated ground objects are removed. The topology is correct if this graph is isomorphic to the graph of the ground-truth simulation model under a categorical match of the node types. A binary isomorphism check is used rather than a graph distance, because flipped kinematic chains or several unconnected subsystems admit more than one valid node correspondence. The search is aborted after 60\,s, and ground-truth graphs beyond 150 nodes are not attempted. Both cases are flagged as unverified rather than counted as failures.

\emph{Parameter agreement.} Given a topological match, each graph node carries the physical parameters of its Exudyn item, such as position, mass or stiffness. Since several node correspondences can be topologically valid, e.g.\ for a flipped kinematic chain, all of them (up to 1000 correspondences, or 120\,s) are searched for the one with the fewest mismatching values. Each value is compared numerically with relative tolerance $\mathrm{rtol}=10^{-4}$ and absolute tolerance $\mathrm{atol}=10^{-6}$, i.e.\ two values $a$ and $b$ agree if $|a-b| \le \mathrm{atol} + \mathrm{rtol}\max(|a|,|b|)$. If the topology could not be verified, no node correspondence is available and the parameter check is skipped. Such simulation models are judged by executability and trajectory agreement alone. In this benchmark, this applies only to the beam bridge (ID 23).

\emph{Trajectory agreement.} In the order of the checks, this check applies to simulation models whose topology and parameters already agree or could not be verified, and thus additionally confirms that the assembled simulation model reproduces the ground-truth motion. The state coordinates of both simulation models (positions, orientations and their time derivatives) are stacked over all solver time steps, and the solution is correct if the norm of their numerical difference is below $\epsilon=0.1$~\cite{Moeltner2026_CreationEvaluationSelfValidation}. This tolerance is wider than the $10^{-5}$ of MecEng~\cite{Gerstmayr2026_LLMAwarenesMechanics}, because the \acp{VLM} often round angle-dependent coordinates (e.g.\ $0.866$ for $\sin 60^\circ$), and slightly inconsistent initial positions cause a short transient when the solver enforces a distance constraint. As the norm accumulates over all coordinates and output times, such deviations alone exceed $10^{-5}$ for otherwise correct simulation models.

\emph{Overall success} requires all four checks. If the topology could not be verified, it rests on executability and trajectory agreement alone. A generated simulation model that matches the ground-truth trajectory with a wrong structure or wrong parameters, e.g.\ a wrong mass in free fall, therefore does not count as successful. Whether Stage 2a selected all components used in the ground-truth simulation model is reported separately, with surplus components not penalized, as it describes the pipeline rather than the generated simulation model.

\subsection{Experimental setup}\label{sec_experimentalSetup}
The 12 open-weight \acp{VLM} are listed in \mytab{tab:overallSuccess}. All \acp{VLM} and the coder \ac{LLM} run unquantized at bf16 precision, served with vLLM~\cite{kwon2023vllm}, with zero sampling temperature and without extended reasoning, on a single NVIDIA RTX PRO 6000 (96\,GB). Each \ac{VLM} processes the 115 standard sketches in all four pipelines, and the coder \ac{LLM} processes the 115 descriptions for the baseline. Repeating this evaluation with 10 instead of 5 parameter sets per mechanical model, i.e.\ on 230 tasks of which $88\,\%$ are drawn independently of the original ones, reproduces every pipeline mean within $0.6$~\acp{PP}. Two pipelines evaluated on the same tasks differ by about 1~\ac{PP} under this repetition, so differences of a few \acp{PP} between pipelines are meaningful. Gaps between \acp{VLM} below about 5~\acp{PP} are not distinguishable: of the 264 pairs of \acp{VLM} within the four pipelines, none with a gap above $7.8$~\acp{PP} changed order between the two task draws, so \acp{VLM} are compared as groups. This repetition and the comparison with the proprietary \ac{VLM} in Sec.~\ref{sec_results} were run with the original research code. The published code, which contains only the components of the original repository required for this project, reproduces the pipeline means of the main evaluation within $0.4$~\acp{PP}, i.e.\ within this run-to-run variation. Differences between pipelines are tested with a two-sided Wilcoxon signed-rank test over the paired success rates of the 12 \acp{VLM}.

\section{Results}\label{sec_results}
\subsection{Comparison of input modalities}\label{sec_pipelineComparison}
\mytab{tab:overallSuccess} gives the overall success of every \ac{VLM} and pipeline on the standard sketches. With the fixed coder \ac{LLM}, the text-only baseline reaches $94.8\,\%$. Since this coder \ac{LLM} succeeds when the mechanical model is described in words, coding ability does not limit the \emph{IR} pipelines, and any loss is caused by the perception of the sketch. Replacing the description by a sketch lowers the mean success to $78.2\,\%$ with the text alongside (\emph{IR + text}) and to $48.0\,\%$ without it (\emph{IR}).

Within the deployable pair, the \ac{IR} stage doubles overall success, from $23.9\,\%$ (\emph{naive}) to $48.0\,\%$ (\emph{IR}), with a median gain of $17.8$~\acp{PP} and an improvement for 11 of 12 \acp{VLM} ($p=0.003$). The gain is largest for \acp{VLM} that fail end-to-end, e.g.\ Qwen3.5-27B ($1.7\,\%$ to $75.7\,\%$) and Qwen3-VL-32B-Instruct ($30.4\,\%$ to $76.5\,\%$), whereas the two \acp{VLM} with the highest \emph{naive} success, gemma-4-31B-it ($74.8\,\%$) and Qwen3.6-27B ($66.1\,\%$), reach similar values with the \ac{IR} ($67.0\,\%$ and $68.7\,\%$). \emph{Naive} and \emph{IR} compare an end-to-end with a separated architecture, because the \ac{VLM} also writes the code in the \emph{naive} pipeline. When a textual description is present, however, the two architectures do not differ systematically (\emph{naive + text} $70.2\,\%$ vs.\ \emph{IR + text} $78.2\,\%$, median $+1.7$~\acp{PP}, better for 6 of 12 \acp{VLM}, $p=0.38$). The change of coder \ac{LLM} therefore does not drive the gain. The \ac{IR} is a perception aid, which helps when the sketch is the only source of the mechanical model.

A same-model comparison with one proprietary \ac{VLM}, Claude Opus~5\footnote{Anthropic, API model \texttt{claude-opus-5}, accessed August 2026.}, which writes the code itself in both conditions, confirms the gap: on the 115 standard tasks it reaches $100.0\,\%$ from the text alone and $79.1\,\%$ (91/115) from the sketch alone (\emph{naive}). Among the open-weight \acp{VLM}, newer generations outperform larger older ones, e.g.\ Qwen3-VL-8B-Instruct ($56.5\,\%$) compared to Qwen2.5-VL-32B-Instruct ($43.5\,\%$) in the \emph{IR} pipeline.

\begin{table}[t]%
\caption{Overall success (in \%) per \ac{VLM} and pipeline on the standard sketches, sorted by the \emph{IR} pipeline, followed by the mean over all 12 \acp{VLM} and the text-only baseline with the fixed coder \ac{LLM}.}
\label{tab:overallSuccess}
\centering
\begin{tabular*}{\linewidth}{@{\extracolsep{\fill}}lrrrr}
\toprule
& \multicolumn{4}{c}{overall success (\%)} \\
\cmidrule(lr){2-5}
VLM & \raisebox{0.5\normalbaselineskip}{\symbolNaiveText}\ \shortstack{naive\\+ text} & \symbolNaive\ naive & \symbolIR\ IR & \raisebox{0.5\normalbaselineskip}{\symbolIRText}\ \shortstack{IR\\+ text} \\
\midrule
Qwen3-VL-32B-Instruct & 94.8 & 30.4 & 76.5 & 88.7 \\
Qwen3.5-27B & 47.8 & 1.7 & 75.7 & 92.2 \\
Qwen3.6-27B & 99.1 & 66.1 & 68.7 & 86.1 \\
gemma-4-31B-it & 100.0 & 74.8 & 67.0 & 87.0 \\
Qwen3.5-9B & 81.7 & 20.0 & 64.3 & 87.8 \\
Qwen3-VL-8B-Instruct & 68.7 & 6.1 & 56.5 & 93.0 \\
Ministral-3-8B-Instruct-2512 & 78.3 & 15.7 & 44.3 & 75.7 \\
Qwen2.5-VL-32B-Instruct & 87.0 & 38.3 & 43.5 & 70.4 \\
Ministral-3-14B-Instruct-2512 & 67.8 & 18.3 & 37.4 & 81.7 \\
gemma-3-27b-it & 85.2 & 14.8 & 24.3 & 77.4 \\
Qwen2.5-VL-7B-Instruct & 31.3 & 0.9 & 17.4 & 67.8 \\
gemma-3-4b-it & 0.9 & 0.0 & 0.9 & 30.4 \\
\midrule
\textit{mean} & 70.2 & 23.9 & 48.0 & 78.2 \\
\midrule
\multicolumn{5}{l}{\textit{\symbolBaseline\ text-only baseline (no VLM): 94.8\%}} \\
\bottomrule
\end{tabular*}

\end{table}

\subsection{Ablation of the intermediate representation}\label{sec_irAblation}
\mytab{tab:irAblation} separates the two design choices of the \ac{IR}. Removing the element pre-selection of Stage 1a lowers overall success for 11 of 12 \acp{VLM} with the JSON \ac{IR} and for 9 of 12 with the free-text \ac{IR}, and the penalty is about twice as large for JSON. Without pre-selection, free text beats JSON by a median of $17.0$~\acp{PP} (11 of 12 \acp{VLM}, $p=0.001$). With pre-selection, the gap shrinks to $6.5$~\acp{PP}, still in favour of free text for every \ac{VLM} that differs (10 better, 2 tied, $p=0.002$). The pre-selection thus absorbs most of the cost of imposing a schema. This reproduces, in a sketch-to-simulation setting, the format-restriction effect reported by~\cite{tam2024letMeSpeakFreely}, which~\cite{fan2026capacityNotFormat} attributes to the limited capacity of the \acp{LLM}. The benefit of the \ac{IR} stage therefore traces to the closed-vocabulary pre-selection, which acts as a perception checklist, rather than to the JSON structure.

\begin{table}[t]%
\caption{Overall success (in \%) on the 115 standard sketches, ablating the \ac{IR} format (JSON vs.\ free text) and the Stage-1a element pre-selection (tags). The JSON column with tags equals the \emph{IR} column of \mytab{tab:overallSuccess}.}
\label{tab:irAblation}
\centering
{\small
\begin{tabular*}{\linewidth}{@{\extracolsep{\fill}}lcccc}
\toprule
 & \multicolumn{2}{c}{with tags (Stage 1a)} & \multicolumn{2}{c}{no tags (no Stage 1a)} \\
\cmidrule(lr){2-3}\cmidrule(lr){4-5}
VLM & JSON \ac{IR} & text \ac{IR} & JSON \ac{IR} & text \ac{IR} \\
\midrule
Qwen3-VL-32B-Instruct & 76.5 & 84.3 & 37.4 & 53.9 \\
Qwen3.5-27B & 75.7 & 81.7 & 42.6 & 60.0 \\
Qwen3.6-27B & 68.7 & 83.5 & 35.7 & 68.7 \\
gemma-4-31B-it & 67.0 & 73.9 & 41.7 & 59.1 \\
Qwen3.5-9B & 64.3 & 64.3 & 37.4 & 47.8 \\
Qwen3-VL-8B-Instruct & 56.5 & 64.3 & 36.5 & 40.0 \\
Ministral-3-8B-Instruct-2512 & 44.3 & 51.3 & 11.3 & 40.0 \\
Qwen2.5-VL-32B-Instruct & 43.5 & 44.3 & 13.9 & 42.6 \\
Ministral-3-14B-Instruct-2512 & 37.4 & 60.0 & 25.2 & 41.7 \\
gemma-3-27b-it & 24.3 & 29.6 & 0.9 & 34.8 \\
Qwen2.5-VL-7B-Instruct & 17.4 & 18.3 & 9.6 & 23.5 \\
gemma-3-4b-it & 0.9 & 0.9 & 3.5 & 2.6 \\
\bottomrule
\end{tabular*}
}
\end{table}

\subsection{Robustness to scaling and rotation}\label{sec_variantRobustness}
\begin{figure}[t]%
\centering
\includegraphics[width=\columnwidth]{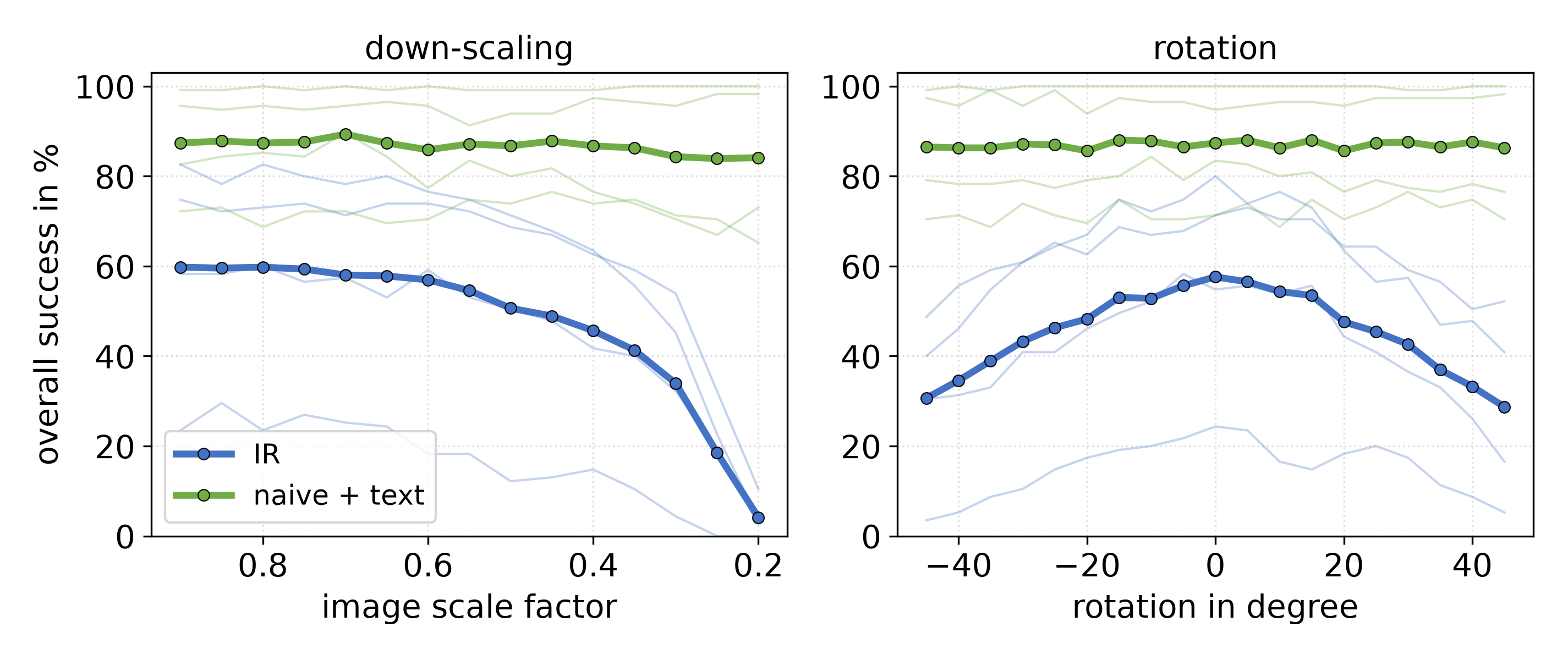}%
\caption{Overall success under down-scaling (left) and rotation (right) for \emph{IR} (sketch only) and \emph{naive + text} (sketch and description). Thick lines show the mean over four \acp{VLM}, thin lines the individual \acp{VLM}.}%
\label{fig:degradationCurves}%
\end{figure}

\fig{fig:degradationCurves} shows the overall success of \emph{IR} and \emph{naive + text} under down-scaling and rotation. For \emph{IR}, success stays near $59\,\%$ from $0.9\times$ to $0.7\times$, falls measurably below $0.6\times$, and reaches $33.9\,\%$ at $0.3\times$, $18.5\,\%$ at $0.25\times$ and $4.1\,\%$ at $0.2\times$ (between 0 and $10.4\,\%$ for the individual \acp{VLM}), where the sketch measures $218\times146$ pixels and its annotations are no longer legible. Rotation degrades \emph{IR} symmetrically beyond $\pm10^\circ$, from $57.6\,\%$ upright to $30.7\,\%$ at $-45^\circ$ and $28.7\,\%$ at $+45^\circ$ (individual \acp{VLM}: $3.5$--$48.7\,\%$ and $5.2$--$52.2\,\%$), and matched angles agree within $2$~\acp{PP} in the mean. \emph{Naive + text} is unaffected by both and stays between $83.9\,\%$ and $89.3\,\%$ under scaling and between $85.7\,\%$ and $88.0\,\%$ under rotation. Perception of the drawing is thus load-bearing only where the drawing is the sole source of the mechanical model.

\subsection{Hand-drawn sketches}\label{sec_handDrawnResults}
\mytab{tab:handDrawnResults} compares the generated and the hand-drawn sketches of the 8 mechanical models, evaluated with Qwen3-VL-32B-Instruct. For \emph{IR}, the hand-drawn sets A, B, and C yield 8, 6 and 7 of 8 successful simulation models, against 7 of 8 for the generated sketches, so no systematic loss from genuine drawings is visible at this sample size. The pipelines that also receive the description solve 7 or 8 of 8 in every set, and \emph{naive} solves 2 of 8 in every set, so the gap between the two deployable pipelines persists on hand-drawn input. One failure recurs in all hand-drawn sets, where \emph{naive + text} misreads the topology of the two mass points with distances (ID 6), while the generated sketch succeeds. Since the three sets share a layout, a cause specific to that layout cannot be excluded.

\begin{table}[t]%
\caption{Successful simulation models out of 8 for the 8 hand-drawn mechanical models (one parameter set each, Qwen3-VL-32B-Instruct): generated sketches against the three hand-drawn sets A, B and C.}
\label{tab:handDrawnResults}
\centering
\begin{tabular*}{\linewidth}{@{\extracolsep{\fill}}lcccc}
\toprule
pipeline & generated sketches & set A & set B & set C \\
\midrule
naive & 2/8 & 2/8 & 2/8 & 2/8 \\
naive + text & 8/8 & 7/8 & 7/8 & 7/8 \\
IR & 7/8 & 8/8 & 6/8 & 7/8 \\
IR + text & 8/8 & 7/8 & 8/8 & 8/8 \\
\bottomrule
\end{tabular*}

\end{table}

\subsection{Origin of the failures}\label{sec_resultserrors}
\begin{figure}[t]%
\centering
\includegraphics[width=\columnwidth]{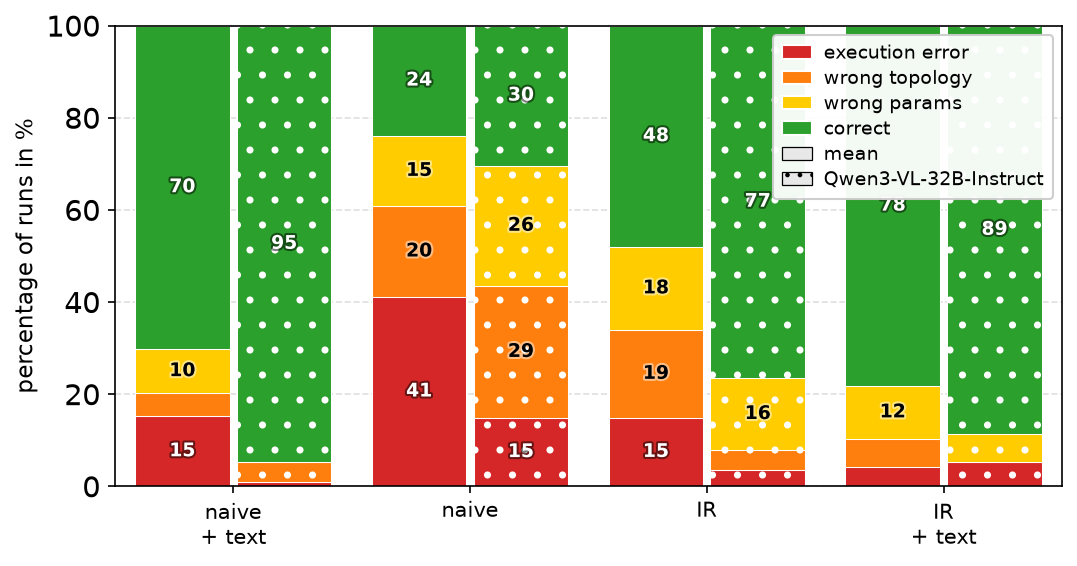}%
\caption{Share of runs per error category and pipeline. For each pipeline, the left bar is the mean over all \acp{VLM} and the right, dotted bar the best \ac{VLM} (Qwen3-VL-32B-Instruct).}%
\label{fig_errorClassification}%
\end{figure}

Each run is assigned to exactly one of four categories, checked in this order: execution error, wrong topology, wrong parameters or correct (\fig{fig_errorClassification}). The categories follow the checks of Sec.~\ref{sec_evaluation}, so the share of correct runs is the overall success. An execution error is broader than a failed executability check, as it also includes runs that execute but yield no comparable solution, e.g.\ a diverging or timed-out solver run. In the \emph{naive} pipeline, execution errors form the largest share, whereas in the \emph{IR} pipeline wrong topology and wrong parameters dominate. Averaged over the \acp{VLM} in the \emph{IR} pipeline, component selection succeeds in $98.1\,\%$ and execution in $91.6\,\%$ of the runs, but the correct topology is reached in only $68.3\,\%$ and correct parameters in $48.1\,\%$. The largest losses therefore arise in reading the structure and the values from the sketch. Once topology and parameters are correct, the trajectory matches as well. 

The only trajectory mismatches with correct topology and parameters, in less than $0.2\,\%$ of all runs, are correct simulation models of the springs bridge (ID 20) whose mass points are numbered in a different order than in the ground truth, which the coordinate-wise trajectory comparison does not map. These simulation codes are conservatively counted as failures.

Infrastructure causes are negligible, as only about $0.1\,\%$ of all runs hit a timeout and none exceeded the context window, so the execution errors are genuine code defects. Comprehension of the sketch, not the simulation of the physics, is the main bottleneck.

\section{Discussion and limitations}\label{sec_discussion}
The results locate the bottleneck of sketch-based model generation in perception rather than in coding or simulation. With a fixed coder \ac{LLM}, the loss from text to sketch is caused by reading the drawing, and once topology and parameters are read correctly, the trajectory follows. The two-stage pipeline targets exactly this step, and its benefit comes from a lightweight closed-vocabulary pre-selection that requires no fine-tuning.

Several design choices bound the scope of these findings. The main benchmark uses generated sketches, which enables systematic variation and verification against ground truth. The hand-drawn study (8 mechanical models in three sets, of which set A was used during development) indicates transfer to genuine drawings, and a larger, independently drawn set would strengthen it. The robustness test covers four \acp{VLM} that span the performance range. The fixed coder \ac{LLM} and the absence of extended reasoning isolate perception and keep inference cost low, as extended reasoning increased inference time by more than a factor of ten for some \acp{LLM} in MecEng~\cite{Gerstmayr2026_LLMAwarenesMechanics}. The proprietary \ac{VLM} was evaluated in the text-only and \emph{naive} conditions, so whether the \ac{IR} also helps proprietary \acp{VLM} remains to be tested. The pipeline runs its stages in a fixed order, and the \acp{LLM} take no decisions about the workflow itself. This keeps each stage evaluable on its own and the comparison of input modalities controlled, but errors are detected not corrected. Agentic approaches, in which \acp{LLM} plan, execute and revise their own steps, as MechAgents~\cite{ni2024mechagents} for mechanics problems or MetaOpenFOAM~\cite{chen2024metaopenfoam} for computational fluid dynamics, could close this loop, e.g.\ by re-reading the sketch after a failed execution.

\section{Conclusions}\label{sec_ConclusionAndOutlook}
Open-weight \acp{VLM} can generate executable multibody simulation models from sketches. With code generation held fixed, replacing the textual description of a mechanical model by a sketch lowers overall success from $94.8\,\%$ to $48.0\,\%$, and to $78.2\,\%$ when the description accompanies the sketch. Separating perception from code generation doubles sketch-only success compared to end-to-end generation by the \ac{VLM} ($48.0\,\%$ vs.\ $23.9\,\%$), and the gain traces to the closed-vocabulary element pre-selection rather than to the output format. The perception is stable down to $0.7\times$ scaling and within $\pm10^\circ$ rotation, degrades beyond, down to $4.1\,\%$ at $0.2\times$ and about $30\,\%$ at $\pm45^\circ$, and transfers to hand-drawn sketches. Most failures arise in reading structure and values from the drawing, while the simulated trajectory follows once these are correct. Being code-agnostic, the \ac{IR} is in principle transferable to other multibody codes, including commercial tools, which is not evaluated here. Future work could embed the pipeline in an agentic workflow, with a critic agent that checks results and triggers re-modeling~\cite{ni2024mechagents}, possibly coupled to learning from execution feedback~\cite{gehring2025_RLEF}, and extend the sketch input to the flexible multibody tasks of MecEng~\cite{Gerstmayr2026_LLMAwarenesMechanics}.

\section*{CRediT authorship contribution statement}
\noindent\textbf{Tobias Möltner:} Methodology, Software, Investigation, Visualization, Validation, Writing -- original draft, Writing -- review \& editing.
\textbf{Peter Manzl:} Methodology, Software, Investigation, Visualization, Validation, Writing -- original draft, Writing -- review \& editing.
\textbf{Sebastian Weyrer:} Methodology, Software, Writing -- original draft, Writing -- review \& editing.
\textbf{Michael Pieber:} Conceptualization, Methodology, Software, Validation, Writing -- original draft, Writing -- review \& editing.

\section*{Declaration of competing interest}
\noindent The authors declare that they have no known competing financial interests or personal relationships that could have appeared to influence the work reported in this paper.

\section*{Funding}
\noindent This research did not receive any specific grant from funding agencies in the public, commercial, or not-for-profit sectors.

\section*{Data availability}
\noindent The code, prompts, and sketches to reproduce the results are available at \url{https://github.com/tmoeltnerUIBK/sketch2sim.git}.

\section*{Declaration of generative AI and AI-assisted technologies in the manuscript preparation process}
\noindent During the preparation of this work the authors used Anthropic's Claude Opus and OpenAI's GPT-6 in order to improve language, check spelling and translate. After using this tool/service, the authors reviewed and edited the content as needed and take full responsibility for the content of the published article.

\section*{Acknowledgments}
\noindent The authors thank Fabian Schuster for contributing one of the three independently drawn sets of hand-drawn sketches used in Sec.~\ref{sec_handDrawnResults} (set C).

\renewcommand{\refname}{References\protect\footnote{For references with more than ten contributing authors, only the first ten authors are listed, followed by ``et al.'', to maintain readability and brevity in the list of references.}}
\bibliographystyle{elsarticle-num}
\bibliography{bibliography}

@article{2020_Brown_languageModelsAreFew-ShotLearners,
  title={{L}anguage {M}odels are {F}ew-{S}hot {L}earners},
  author={Brown, Tom and Mann, Benjamin and Ryder, Nick and Subbiah, Melanie and Kaplan, Jared D and Dhariwal, Prafulla and Neelakantan, Arvind and Shyam, Pranav and Sastry, Girish and Askell, Amanda and others},
  journal={Advances in Neural Information Processing Systems},
  volume={34},
  pages={1877--1901},
  year={2020}
}

@misc{2019_Radford_languageModelsUnsupervisedMultitaskLearners,
  title={{L}anguage {M}odels are {U}nsupervised {M}ultitask {L}earners},
  author={{OpenAI}},
  year = {2019},
  howpublished = {\url{https://openai.com/index/better-language-models/} (accessed on November 29, 2026)}
}

@article{2020_Kaplan_scalingLawsForNeuralLanguageModels,
  title={{S}caling {L}aws for {N}eural {L}anguage {M}odels},
  author={Kaplan, Jared and McCandlish, Sam and Henighan, Tom and Brown, Tom B and Chess, Benjamin and Child, Rewon and Gray, Scott and Radford, Alec and Wu, Jeffrey and Amodei, Dario},
  journal={arXiv preprint},
  year={2020},
	doi={10.48550/arXiv.2001.08361}
}

@article{2017_Vaswani_attentionIsAllYouNeed,
  title={Attention is all you need},
  author={Vaswani, Ashish and Shazeer, Noam and Parmar, Niki and Uszkoreit, Jakob and Jones, Llion and Gomez, Aidan N and Kaiser, {\L}ukasz and Polosukhin, Illia},
  journal={Advances in Neural Information Processing Systems},
  volume={30},
  year={2017}
}

@inproceedings{2019_Devlin_BERT_PreTrainingDeepBidirectionalTransformersForLanguagaeUnderstanding,
    title={{BERT}: Pre-training of {D}eep {B}idirectional {T}ransformers for {L}anguage {U}nderstanding},
    author={Devlin, Jacob and Chang, Ming-Wei and Lee, Kenton and Toutanova, Kristina},
    booktitle={Proceedings of the 2019 Conference of the North {A}merican Chapter of the Association for Computational Linguistics: Human Language Technologies, Volume 1 (Long and Short Papers)},
    year={2019},
    publisher={Association for Computational Linguistics},
    doi={10.18653/v1/N19-1423},
    pages={4171--4186}
}

@inproceedings{Lewis_2020_BART,
    title={{BART}: Denoising {S}equence-to-{S}equence {P}re-training for {N}atural {L}anguage {G}eneration, {T}ranslation, and {C}omprehension},
    author={Lewis, Mike and Liu, Yinhan and Goyal, Naman and Ghazvininejad, Marjan and Mohamed, Abdelrahman and Levy, Omer and Stoyanov, Veselin and Zettlemoyer, Luke},
    booktitle={Proceedings of the 58th Annual Meeting of the Association for Computational Linguistics},
    year={2020},
    publisher={Association for Computational Linguistics},
    doi={10.18653/v1/2020.acl-main.703},
    pages={7871--7880}
}

@article{Exudyn2024,
	author={Gerstmayr, Johannes},
	title={{E}xudyn -- a {C}++-based {P}ython package for flexible multibody systems},
	journal={Multibody System Dynamics},
	year={2024},
	volume={60},
	pages={533--561},
  doi={10.1007/s11044-023-09937-1}
}

@article{TasoraNegrutChrono2013,
author = {Mazhar, Hammad and Heyn, Toby and Pazouki, Arman and Melanz, Dan and Seidl, Andrew and Bartholomew, Aaron and Tasora, Alessandro and Negrut, Dan},
year = {2013},
month = {02},
pages = {49--64},
title = {{CHRONO: A} parallel multi-physics library for rigid-body, flexible-body, and fluid dynamics},
volume = {4},
journal = {Mechanical Sciences},
doi = {10.5194/ms-4-49-2013}
}

@misc{2018_radford_improvingLanguageUnderstandingGPT,
	title={Improving {L}anguage {U}nderstanding by {G}enerative {P}re-{T}raining},
  author={{OpenAI}},
  year={2018},
  howpublished = {\url{https://openai.com/index/language-unsupervised/} (accessed on November 29, 2026)}
}

@inproceedings{kwon2023vllm,
    title={{E}fficient {M}emory {M}anagement for {L}arge {L}anguage {M}odel {S}erving with {P}aged{A}ttention},
    author={Kwon, Woosuk and Li, Zhuohan and Zhuang, Siyuan and Sheng, Ying and Zheng, Lianmin and Yu, Cody Hao and Gonzalez, Joseph E. and Zhang, Hao and Stoica, Ion},
    booktitle={Proceedings of the 29th Symposium on Operating Systems Principles (SOSP)},
    year={2023},
    pages={611--626},
    doi={10.1145/3600006.3613165}
}

@article{Hui2024qwen2_5Coder,
  title={{Q}wen2.5-{C}oder {T}echnical {R}eport},
  author={Hui, Binyuan and Yang, Jian and Cui, Zeyu and Yang, Jiaxi and Liu, Dayiheng and Zhang, Lei and Liu, Tianyu and Zhang, Jiajun and Yu, Bowen and Lu, Keming and others},
  journal={arXiv preprint},
  year={2024},
	doi={10.48550/arXiv.2409.12186}
}

@article{Guo2024deepseekCoder_LLMMeetsProgramming,
  title={{D}eepSeek-{C}oder: {W}hen the {L}arge {L}anguage {M}odel {M}eets {P}rogramming--{T}he {R}ise of {C}ode {I}ntelligence},
  author={Guo, Daya and Zhu, Qihao and Yang, Dejian and Xie, Zhenda and Dong, Kai and Zhang, Wentao and Chen, Guanting and Bi, Xiao and Wu, Yu and Li, YK and others},
  journal={arXiv preprint},
  year={2024}, 
	doi={10.48550/arXiv.2401.14196}
}

@inproceedings{zhuo2025bigcodebench,
  title={{B}ig{C}ode{B}ench: {B}enchmarking {C}ode {G}eneration with {D}iverse {F}unction {C}alls and {C}omplex {I}nstructions},
  author={Zhuo, Terry Yue and Vu, Minh Chien and Chim, Jenny and Hu, Han and Yu, Wenhao and Widyasari, Ratnadira and Yusuf, Imam Nur Bani and Zhan, Haolan and He, Junda and Paul, Indraneil and others},
  booktitle={Proceedings of the International Conference on Learning Representations (ICLR)},
	pages={66602--66656},
  year={2025}
}

@article{wang2024qwen2VL,
  title={{Q}wen2-{VL}: {E}nhancing {V}ision-{L}anguage {M}odel's {P}erception of the {W}orld at {A}ny {R}esolution},
  author={Wang, Peng and Bai, Shuai and Tan, Sinan and Wang, Shijie and Fan, Zhihao and Bai, Jinze and Chen, Keqin and Liu, Xuejing and Wang, Jialin and Ge, Wenbin and others},
  journal={arXiv preprint},
  year={2024}, 
	doi={10.48550/arXiv.2409.12191}
}

@article{Moeltner2026_CreationEvaluationSelfValidation,
title = {Creation, evaluation and self-validation of simulation models with large language models},
journal = {Neurocomputing},
volume = {663},
pages = {132030},
year = {2026},
issn = {0925-2312},
doi = {10.1016/j.neucom.2025.132030},
author = {Tobias Möltner and Peter Manzl and Michael Pieber and Johannes Gerstmayr},
}

@article{gerstmayr2024multibodyModelsGenerated,
  title={Multibody {M}odels {G}enerated from {N}atural {L}anguage},
  author={Gerstmayr, Johannes and Manzl, Peter and Pieber, Michael},
  journal={Multibody System Dynamics},
  volume={62},
  number={2},
  pages={249--271},
  year={2024},
  publisher={Springer},
	doi={10.1007/s11044-023-09962-0}
}

@article{ouyang2022trainingLLMsRLHF,
  title={Training language models to follow instructions with human feedback},
  author={Ouyang, Long and Wu, Jeffrey and Jiang, Xu and Almeida, Diogo and Wainwright, Carroll and Mishkin, Pamela and Zhang, Chong and Agarwal, Sandhini and Slama, Katarina and Ray, Alex and others},
  journal={Advances in Neural Information Processing Systems},
  volume={35},
  pages={27730--27744},
  year={2022}, 
	doi={10.52202/068431-2011}
}

@inproceedings{bender2021dangers,
  title={{O}n the {D}angers of {S}tochastic {P}arrots: {C}an {L}anguage {M}odels be {T}oo {B}ig?},
  author={Bender, Emily M and Gebru, Timnit and McMillan-Major, Angelina and Shmitchell, Shmargaret},
  booktitle={Proceedings of the 2021 ACM conference on fairness, accountability, and transparency},
  pages={610--623},
  year={2021}, 
	doi={10.1145/3442188.3445922}	
}

@article{hendrycks2020MMLU,
  title={{M}easuring {M}assive {M}ultitask {L}anguage {U}nderstanding},
  author={Hendrycks, Dan and Burns, Collin and Basart, Steven and Zou, Andy and Mazeika, Mantas and Song, Dawn and Steinhardt, Jacob},
  journal={arXiv preprint},
  year={2021},
	doi={10.48550/arXiv.2009.03300}
}

@inproceedings{redmon_2016_YOLO_original,
  title={{Y}ou {O}nly {L}ook {O}nce: {U}nified, {R}eal-{T}ime {O}bject {D}etection},
  author={Redmon, Joseph and Divvala, Santosh and Girshick, Ross and Farhadi, Ali},
  booktitle={Proceedings of the IEEE conference on computer vision and pattern recognition},
  pages={779--788},
  year={2016}, 
	doi={10.1109/CVPR.2016.91}
}

@article{cai2021PINNS_heat,
  title={{P}hysics-{I}nformed {N}eural {N}etworks for {H}eat {T}ransfer {P}roblems},
  author={Cai, Shengze and Wang, Zhicheng and Wang, Sifan and Perdikaris, Paris and Karniadakis, George Em},
  journal={Journal of Heat Transfer},
  volume={143},
  number={6},
  pages={060801},
  year={2021},
  publisher={American Society of Mechanical Engineers},
	doi={10.1115/1.4050542}
}

@article{manzl2025SLIDE,
  title={{SLIDE}: A machine-learning based method for forced dynamic response estimation of multibody systems},
  author={Manzl, Peter and Humer, Alexander and Khadim, Qasim and Gerstmayr, Johannes},
  journal={Mechanics Based Design of Structures and Machines},
	volume = {54},
	number = {1},
	pages = {2559325},
	year = {2026},
  publisher={Taylor \& Francis},
	doi={10.1080/15397734.2025.2559325}
}

@article{wang2025Chronollm,
  title={Chrono{LLM}: Customizing language models for physics-based simulation code generation},
  author={Wang, Jingquan and Negrut, Andrew and Zhang, Harry and Slaton, Khailanii and Wang, Shu and Serban, Radu and Wu, Jinlong and Negrut, Dan},
  journal={Multibody System Dynamics},
	volume={67},
	pages={895--939},
  year={2026}, 
	publisher={Springer},
	doi={10.1007/s11044-026-10152-x} 
}

@article{wang2025mbd,
  title={{MBD-NODE}: physics-informed data-driven modeling and simulation of constrained multibody systems},
  author={Wang, Jingquan and Wang, Shu and Unjhawala, Huzaifa Mustafa and Wu, Jinlong and Negrut, Dan},
  journal={Multibody System Dynamics},
  volume={64},
  number={3},
  pages={543--585},
  year={2025},
  publisher={Springer},
	doi={10.1007/s11044-024-10012-6}
}

@article{klein2022PANN,
  title={Finite electro-elasticity with physics-augmented neural networks},
  author={Klein, Dominik K and Ortigosa, Rogelio and Mart{\'\i}nez-Frutos, Jes{\'u}s and Weeger, Oliver},
  journal={Computer Methods in Applied Mechanics and Engineering},
  volume={400},
  pages={115501},
  year={2022},
  publisher={Elsevier}, 
	doi={10.1016/j.cma.2022.115501}
}

@article{chen2018neuralODEs,
  title={Neural {O}rdinary {D}ifferential {E}quations},
  author={Chen, Ricky T Q and Rubanova, Yulia and Bettencourt, Jesse and Duvenaud, David K},
  journal={arXiv preprint},
  year={2018}, 
	doi={10.48550/arXiv.1806.07366}
}

@article{rai2020explainableAI,
  title={{Explainable} {AI}: {F}rom black box to glass box},
  author={Rai, Arun},
  journal={Journal of the Academy of Marketing Science},
  volume={48},
  pages={137--141},
  year={2020},
  publisher={Springer}, 
	doi={10.1007/s11747-019-00710-5}
}

@inproceedings{yang2017investigatingGreyBoxModeling,
  title={Investigating {G}rey-{B}ox {M}odeling for {P}redictive {A}nalytics in {S}mart {M}anufacturing},
  author={Yang, Zhuo and Eddy, Douglas and Krishnamurty, Sundar and Grosse, Ian and Denno, Peter and Lu, Yan and Witherell, Paul},
  booktitle={Proceedings of the ASME 2017 International Design Engineering Technical Conferences and Computers and Information in Engineering Conference},
  volume={2B: 43rd Design Automation Conference},
  pages={V02BT03A024},
  year={2017},
  organization={American Society of Mechanical Engineers}, 
	doi={10.1115/DETC2017-67794}
}

@article{teamQwen2023qwenVL,
  title={{Q}wen-{VL}: {A} {V}ersatile {V}ision-{L}anguage {M}odel for {U}nderstanding, {L}ocalization, {T}ext {R}eading, and {B}eyond},
  author={Bai, Jinze and Bai, Shuai and Yang, Shusheng and Wang, Shijie and Tan, Sinan and Wang, Peng and Lin, Junyang and Zhou, Chang and Zhou, Jingren},
  journal={arXiv preprint},
  year={2023},
	doi={10.48550/arXiv.2308.12966}
}

@article{yang2025qwen3,
  title={Qwen3 {T}echnical {R}eport},
  author={Yang, An and Li, Anfeng and Yang, Baosong and Zhang, Beichen and Hui, Binyuan and Zheng, Bo and Yu, Bowen and Gao, Chang and Huang, Chengen and Lv, Chenxu and others},
  journal={arXiv preprint},
  year={2025},
	doi={10.48550/arXiv.2505.09388}
}

@article{bai2025qwen3VL,
  title={{Q}wen3-{VL} {T}echnical {R}eport},
  author={Bai, Shuai and Cai, Yuxuan and Chen, Ruizhe and Chen, Keqin and Chen, Xionghui and Cheng, Zesen and Deng, Lianghao and Ding, Wei and Gao, Chang and Ge, Chunjiang and others},
  journal={arXiv preprint},
  year={2025}, 
	doi={10.48550/arXiv.2511.21631}
}

@inproceedings{qian2023limitationsofLLMsArithmetics,
  title={{L}imitations of {L}anguage {M}odels in {A}rithmetic and {S}ymbolic {I}nduction},
  author={Qian, Jing and Wang, Hong and Li, Zekun and Li, Shiyang and Yan, Xifeng},
  booktitle={Proceedings of the 61st Annual Meeting of the Association for Computational Linguistics (Volume 1: Long Papers)},
  pages={9285--9298},
  year={2023}, 
	doi={10.18653/v1/2023.acl-long.516}
}

@article{dosovitskiy2020VisionTransformer,
  title={{A}n {I}mage is {W}orth 16x16 {W}ords: {T}ransformers for {I}mage {R}ecognition at {S}cale},
  author={Dosovitskiy, Alexey and Beyer, Lucas and Kolesnikov, Alexander and Weissenborn, Dirk and Zhai, Xiaohua and Unterthiner, Thomas and Dehghani, Mostafa and Minderer, Matthias and Heigold, Georg and Gelly, Sylvain and others},
  journal={arXiv preprint},
  year={2021}, 
  doi={10.48550/arXiv.2010.11929},
}

@InProceedings{Caron_2021_SelfSupervisedViT,
    author    = {Caron, Mathilde and Touvron, Hugo and Misra, Ishan and J\'egou, Herv\'e and Mairal, Julien and Bojanowski, Piotr and Joulin, Armand},
    title     = {{E}merging {P}roperties in {S}elf-{S}upervised {V}ision {T}ransformers},
    booktitle = {Proceedings of the IEEE/CVF International Conference on Computer Vision (ICCV)},
    month     = {October},
    year      = {2021},
    pages     = {9650-9660}, 
		doi={10.1109/ICCV48922.2021.00951}
}

@inproceedings{He2022maskedAutoencoders_ViT,
  title={{M}asked {A}utoencoders are {S}calable {V}ision {L}earners},
  author={He, Kaiming and Chen, Xinlei and Xie, Saining and Li, Yanghao and Doll{\'a}r, Piotr and Girshick, Ross},
  booktitle={Proceedings of the IEEE/CVF Conference on Computer Vision and Pattern Recognition},
  pages={16000--16009},
  year={2022},
	doi={10.1109/CVPR52688.2022.01553}
}

@inproceedings{radford2021learningTransferableVisualModelsFromNLPSuperv,
  title={{L}earning {T}ransferable {V}isual {M}odels {F}rom {N}atural {L}anguage {S}upervision},
  author={Radford, Alec and Kim, Jong Wook and Hallacy, Chris and Ramesh, Aditya and Goh, Gabriel and Agarwal, Sandhini and Sastry, Girish and Askell, Amanda and Mishkin, Pamela and Clark, Jack and Krueger, Gretchen and Sutskever, Ilya},
  booktitle={Proceedings of the 38th International Conference on Machine Learning (PMLR)},
  pages={8748--8763},
  year={2021},
  volume={139}
}

@misc{openai2023gpt4v,
  title={{GPT}-4{V}(ision) {S}ystem {C}ard},
  author={{OpenAI}},
  year={2023},
  howpublished = {\url{https://openai.com/index/gpt-4v-system-card/} (accessed on November 29, 2026)}
}

@article{Liu2023Llava,
 author = {Liu, Haotian and Li, Chunyuan and Wu, Qingyang and Lee, Yong Jae},
 journal = {Advances in Neural Information Processing Systems},
 pages = {34892--34916},
 title = {{V}isual {I}nstruction {T}uning},
 volume = {36},
 year = {2023}, 
 doi={10.52202/075280-1516}
}

@inproceedings{jimenez2024SWEBench,
  title={{SWE}-bench: {C}an {L}anguage {M}odels {R}esolve {R}eal-world {G}ithub {I}ssues?},
  author={Jimenez, Carlos E and Yang, John and Wettig, Alexander and Yao, Shunyu and Pei, Kexin and Press, Ofir and Narasimhan, Karthik},
  booktitle={Proceedings of the International Conference on Learning Representations (ICLR)},
  pages={54107--54157},
  year={2024}, 
}

@article{wang2024MathVisionBenchmark,
  title={{M}easuring {M}ultimodal {M}athematical {R}easoning with {MATH}-{V}ision {D}ataset},
  author={Wang, Ke and Pan, Junting and Shi, Weikang and Lu, Zimu and Ren, Houxing and Zhou, Aojun and Zhan, Mingjie and Li, Hongsheng},
  journal={Advances in Neural Information Processing Systems},
  volume={37},
  pages={95095--95169},
  year={2024}, 
	doi={10.52202/079017-3014}
}

@inproceedings{Si2025Design2Code,
    title={{D}esign2{C}ode: {B}enchmarking {M}ultimodal {C}ode {G}eneration for {A}utomated {F}ront-{E}nd {E}ngineering},
    author={Si, Chenglei and Zhang, Yanzhe and Li, Ryan and Yang, Zhengyuan and Liu, Ruibo and Yang, Diyi},
    booktitle={Proceedings of the 2025 Conference of the Nations of the Americas Chapter of the Association for Computational Linguistics: Human Language Technologies (Volume 1: Long Papers)},
    year={2025},
		pages={3956--3974},
    publisher={Association for Computational Linguistics},
    doi={10.18653/v1/2025.naacl-long.199}
}

@article{gehring2025_RLEF,
  title={{RLEF}: {G}rounding {C}ode {LLMs} in {E}xecution {F}eedback with {R}einforcement {L}earning},
  author={Gehring, Jonas and Zheng, Kunhao and Copet, Jade and Mella, Vegard and Cohen, Taco and Synnaeve, Gabriel},
  journal={arXiv preprint}, 
	year={2025}, 
  doi={10.48550/arXiv.2410.02089}
}

@article{ni2024mechagents,
  title={Mech{A}gents: Large language model multi-agent collaborations can solve mechanics problems, generate new data, and integrate knowledge},
  author={Ni, Bo and Buehler, Markus J},
  journal={Extreme Mechanics Letters},
  volume={67},
  pages={102131},
  year={2024},
  publisher={Elsevier}, 
	doi={10.1016/j.eml.2024.102131}
}

@article{Singh2025GPT5ModelCard,
  title={{O}penai {GPT}-5 {S}ystem {C}ard},
  author={Singh, Aaditya and Fry, Adam and Perelman, Adam and Tart, Adam and Ganesh, Adi and El-Kishky, Ahmed and McLaughlin, Aidan and Low, Aiden and Ostrow, AJ and Ananthram, Akhila and others},
  journal={arXiv preprint},
  year={2026}, 
	doi={10.48550/arXiv.2601.03267}
}

@article{bubeck2023paperSparksOfAGI,
  title={{S}parks of {A}rtificial {G}eneral {I}ntelligence: {E}arly experiments with {GPT}-4},
  author={Bubeck, S{\'e}bastien and Chandrasekaran, Varun and Eldan, Ronen and Gehrke, Johannes and Horvitz, Eric and Kamar, Ece and Lee, Peter and Lee, Yin Tat and Li, Yuanzhi and Lundberg, Scott and others},
	journal={arXiv preprint},
  year={2023}, 
	doi={10.48550/arXiv.2303.12712}
}

@article{Gerstmayr2026_LLMAwarenesMechanics,
  title={{L}arge {L}anguage {M}odels and their {A}wareness of {M}echanics and {S}patial {G}eometry},
  author={Johannes Gerstmayr and Sebastian Weyrer and Tobias Möltner and Peter Manzl and Michael Pieber},
	journal={arXiv preprint},
  year={2026},
	doi={10.48550/arXiv.2608.14615}
}

@article{kamath2025gemma3,
  title={{G}emma 3 {T}echnical {R}eport},
  author={Kamath, Aishwarya and Ferret, Johan and Pathak, Shreya and Vieillard, Nino and Merhej, Ramona and Perrin, Sarah and Matejovicova, Tatiana and Ram{\'e}, Alexandre and Rivi{\`e}re, Morgane and Rouillard, Louis and others},
  journal={arXiv preprint},
  year={2025},
  doi={10.48550/arXiv.2503.19786}
}

@article{gemmateam2026gemma4,
  title={{G}emma 4 {T}echnical {R}eport},
  author={El Abd, Sherif and Aggarwal, Vaibhav and Algayres, Robin and Andreev, Alek and Bachem, Olivier and Ballantyne, Ian and Brick, Cormac and C\u{a}rbune, Victor and Casbon, Michelle and Chaturvedi, Mayank and others},
  journal={arXiv preprint},
  year={2026},
  doi={10.48550/arXiv.2607.02770}
}

@article{bai2025qwen25vl,
  title={{Q}wen2.5-{VL} {T}echnical {R}eport},
  author={Bai, Shuai and Chen, Keqin and Liu, Xuejing and Wang, Jialin and Ge, Wenbin and Song, Sibo and Dang, Kai and Wang, Peng and Wang, Shijie and Tang, Jun and others},
  journal={arXiv preprint},
  year={2025},
  doi={10.48550/arXiv.2502.13923}
}

@article{mistralai2025ministral3,
  author={H. Liu, Alexander and Khandelswal, Kartik and Subramanian, Sandeep and Jouault, Victor and Rastogi, Abhinav and Sad\'e, Adrien and Jeffares, Alan and Jiang, Albert and Cahill, Alexandre and Sablayrolles, Alexandre and others},
  title={{M}inistral 3 14{B} {M}odel {C}ard},
  journal={arXiv preprint},
  year={2026},
  doi={10.48550/arXiv.2601.08584}
}

@article{Ezemba2026OpenSeeSimE,
  author={Ezemba, Jessica and Pohl, Jason and Tucker, Conrad and McComb, Christopher},
  title={A large-scale benchmark to assess vision-language model question answering capabilities in engineering simulations},
  year={2026},
  journal={Communications Engineering},
  doi={10.1038/s44172-026-00786-2}
}

@inproceedings{Zeng2025LEGOpuzzles,
author={Tang, Kexian and Gao, Junyao and Zeng, Yanhong and Duan, Haodong and Sun, Yanan and Xing, Zhening and Liu, Wenran and Chen, Kai and Lyu, Kaifeng},
title={{LEGO}-{P}uzzles: {H}ow {G}ood {A}re {MLLM}s at {M}ulti-step {S}patial {R}easoning?},
booktitle={Computer Vision -- ECCV 2026},
year={2026},
publisher={Springer Nature Switzerland},
pages={1--19},
doi={10.1007/978-3-032-37553-7_1}
}

@article{khan2024text2cad,
  title={Text2{CAD}: {G}enerating {S}equential {CAD} {D}esigns from {B}eginner-to-{E}xpert {L}evel {T}ext {P}rompts},
  author={Khan, Mohammad S and Sinha, Sankalp and Sheikh, Talha U and Stricker, Didier and Ali, Sk A and Afzal, Muhammad Z},
  journal={Advances in Neural Information Processing Systems},
  volume={37},
  pages={7552--7579},
  year={2024},
	doi={10.52202/079017-0242}
}

@article{guan2026cad,
  title={{CAD}-{C}oder: {T}ext-to-{CAD} {G}eneration with {C}hain-of-{T}hought and {G}eometric {R}eward},
  author={Guan, Yandong and Wang, Xilin and Xing, Ximing and Zhang, Jing and Xu, Dong and Yu, Qian},
  journal={Advances in Neural Information Processing Systems},
  volume={38},
  pages={59765--59789},
  year={2025},
	doi={10.52202/085713-1999}
}

@article{molinari2026engiai,
  title={Engi{AI}: {A} {M}ulti-{A}gent {F}ramework and {B}enchmark {S}uite for {LLM}-{D}riven {E}ngineering {D}esign},
  author={Molinari, Gioele and Felten, Florian and Massoudi, Soheyl and Fuge, Mark},
  journal={arXiv preprint},
  year={2026},
	doi={10.48550/arXiv.2605.19743}
}

@article{Zhou2025EngiBench,
  author={Zhou, Xiyuan and Wang, Xinlei and He, Yirui and Wu, Yang and Zou, Ruixi and Cheng, Yuheng and Xie, Yulu and Liu, Wenxuan and Zhao, Huan and Xu, Yan and others},
  title={{EngiBench}: {A} {B}enchmark for {E}valuating {L}arge {L}anguage {M}odels on {E}ngineering {P}roblem {S}olving},
  year={2025},
  journal={arXiv preprint},
  doi={10.48550/arXiv.2509.17677}
}

@article{Abdullaeva2026NoReGeo,
  author  = {Abdullaeva, Irina and Vasiliuk, Anton and Goncharova, Elizaveta and Rahmatullaev, Temurbek and Ivan, Zagorulko and Kurkin, Maxim and Kuznetsov, Andrey},
  title   = {{NoReGeo}: {N}on-{R}easoning {G}eometry {B}enchmark},
  journal = {Proceedings of the AAAI Conference on Artificial Intelligence},
  volume  = {40},
  number  = {23},
  pages   = {18901--18909},
  year    = {2026},
  doi     = {10.1609/aaai.v40i23.38960}
}

@article{Sharma2023SpatialReasoningLLM,
  author={Sharma, Manasi},
  title={{E}xploring and {I}mproving the {S}patial {R}easoning {A}bilities of {L}arge {L}anguage {M}odels},
  journal={arXiv preprint},
  year={2023},
	doi={10.48550/arXiv.2312.01054}
}

@article{Yamada2024SpatialLLM,
  author={Yamada, Yutaro and Bao, Yihan and Lampinen, Andrew K. and Kasai, Jungo and Yildirim, Ilker},
  title={{E}valuating {S}patial {U}nderstanding of {L}arge {L}anguage {M}odels},
  journal={arXiv preprint},
  year={2024},
	doi={10.48550/arXiv.2310.14540}
}

@article{edwards2024sketch2prototype,
  title={{Sketch2Prototype}: {R}apid {C}onceptual {D}esign {E}xploration and {P}rototyping with {G}enerative {AI}},
  author={Edwards, Kristen M and Man, Brandon and Ahmed, Faez},
  journal={arXiv preprint},
  year={2024},
	doi={10.48550/arXiv.2405.12985}
}

@inproceedings{tam2024letMeSpeakFreely,
    title={{L}et {M}e {S}peak {F}reely? {A} {S}tudy on the {I}mpact of {F}ormat {R}estrictions on {P}erformance of {L}arge {L}anguage {M}odel },
    author={Tam, Zhi Rui and Wu, Cheng-Kuang and Tsai, Yi-Lin and Lin, Chieh-Yen and Lee, Hung-yi and Chen, Yun-Nung},
    booktitle={Proceedings of the 2024 Conference on Empirical Methods in Natural Language Processing: Industry Track},
    year={2024},
    publisher={Association for Computational Linguistics},
    doi={10.18653/v1/2024.emnlp-industry.91},
    pages={1218--1236}
}

@article{fan2026capacityNotFormat,
  title={{C}apacity, {N}ot {F}ormat: {R}ethinking {S}tructured {R}easoning {F}ailures},
  author={Fan, Hengxin},
  journal={arXiv preprint},
	doi={10.48550/arXiv.2606.09410},
  year={2026}
}

@article{wang2026simbench,
  title={{SimBench}: A {F}ramework for {E}valuating and {D}iagnosing {LLM}-{B}ased {D}igital-{T}win {G}eneration for {M}ulti-{P}hysics {S}imulation}, 
  author={Wang, Jingquan and Negrut, Andrew and Wang, Hongyu and Zhang, Harry and Negrut, Dan},
  journal={IEEE Access}, 
  year={2026},
  volume={14},
  pages={61784-61808},
  doi={10.1109/ACCESS.2026.3685519}
}

@article{mudur2025feabench,
  title={{FEABench}: {E}valuating {L}anguage {M}odels on {M}ultiphysics {R}easoning {A}bility},
  author={Mudur, Nayantara and Cui, Hao and Venugopalan, Subhashini and Raccuglia, Paul and Brenner, Michael P. and Norgaard, Peter},
  journal={arXiv preprint},
  year={2025},
  doi={10.48550/arXiv.2504.06260}
}

@article{chen2024metaopenfoam,
  title={{MetaOpenFOAM}: an {LLM}-based multi-agent framework for {CFD}},
  author={Chen, Yuxuan and Zhu, Xu and Zhou, Hua and Ren, Zhuyin},
  journal={arXiv preprint},
  year={2024},
  doi={10.48550/arXiv.2407.21320}
}

\end{document}